\documentclass{article}
\usepackage{spconf,amsmath,epsfig}
\usepackage{amstext,amssymb,amsbsy,amsmath,amscd}
\usepackage{graphics,graphicx,epsfig,subfigure,enumerate,color,pstricks,cancel,rotating,theorem,array,longtable,multirow,multicol}

\newcommand{\vtheta}{{\boldsymbol{\theta}}}
\newcommand{\veta}{{\boldsymbol{\eta}}}
\newcommand{\vmu}{{\boldsymbol{\mu}}}

\newcommand{\vT}{{\mathbf{T}}}

\newcommand{\x}{\mathbf{x}}
\newcommand{\y}{\mathbf{y}}
\newcommand{\Y}{\mathbf{Y}}

\newcommand{\V}{\mathbf{V}}
\newcommand{\U}{\mathcal{U}}
\newcommand{\R}{\mathbb{R}}
\newcommand{\T}{\mathbf{T}}
\newcommand{\N}{\mathbf{N}}

\newcommand{\Om}{\Omega}

\newcommand{\ESmo}{\widehat{\theta_i}_{MO}}
\newcommand{\ESmv}{\widehat{\theta_i}_{ML}}

\newcommand{\be}{\begin{eqnarray}}
\newcommand{\ee}{\end{eqnarray}}

\newcommand{\vetaML}{\hat{\mathbf{\eta}}_{ML}}

\newtheorem{definition}{Definition}
\newtheorem{theorem}{Theorem}
\newtheorem{corollaire}{Corollary}
\title{Statistical region-based active contours with exponential family observations}
\name{Fran\c{c}ois Lecellier$^a$, St\'ephanie Jehan-Besson$^a$, Jalal Fadili$^a$, Gilles Aubert$^b$, Marinette Revenu$^a$} 
\address{\begin{tabular}{c c}
    $^a$ GREYC UMR 6072 CNRS  & $^b$ Laboratoire J.A. Dieudonn\'e UMR 6621 CNRS\\
    ENSICAEN-Universit\'e de Caen &        Universit\'e de Nice Sophia-Antipolis \\
    14050 Caen, France& 06108 Nice, France\end{tabular}}

\begin{document}

\maketitle

\begin{abstract}
In this paper, we focus on statistical region-based active contour models where image features (e.g. intensity) are random variables whose distribution belongs to some parametric family (e.g. exponential) rather than confining ourselves to the special Gaussian case. Using shape derivation tools, our effort focuses on constructing a general expression for the derivative of the energy (with respect to a domain) and derive the corresponding evolution speed. A general result is stated within the framework of multi-parameter exponential family. More particularly, when using Maximum Likelihood estimators, the evolution speed has a closed-form expression that depends simply on the probability density function, while complicating additive terms appear when using other estimators, e.g. moments method. Experimental results on both synthesized and real images demonstrate the applicability of our approach.
\end{abstract}

\section{Introduction}
\label{sec:intro}
In image segmentation, the main issue is to extract one or several regions according to a given criterion. Region based active contours have proven their efficiency for such a task. The evolution equation of the deformable curve is deduced from a functional to minimize that may take benefit of statistical properties of the image features. In this case, the functional depends on the probability density function (pdf) of the image feature within the region. For instance, the minimization of the - log-likelihood of the pdf has been widely used for the segmentation of homogeneous regions in noisy images \cite{Zhu_pami_96,Paragios_jvcir_02,Refregier_ip04}. In these works, the pdf is parametric, i.e. it follows a prespecified law (Gaussian, Rayleigh). The pdf is then indexed by one or more parameters (e.g. mean and variance for Gaussian laws) describing the distribution model. These parameters depend on the region and must be estimated at each evolution step. The estimation may be performed using different methods (moments method, maximum likelihood (ML)). However, to the best of our knowledge, the influence of the estimation method on the computation of the evolution equation has never been investigated.

In this paper, using shape derivative tools \cite{Delfour_book_01,Jehan_ijcv_02,Jehan_siam_03}, our effort focuses on constructing a general expression for the derivative of a functional depending on parametric pdfs, and hence on the corresponding evolution equation. Our contribution is twofold. Firstly, we show that the estimator of the distribution parameters is crucial for the derived speed expression. Secondly, we propose to give general results for the evolution equation within the framework of multi-parameter exponential family and ML estimation theory. Therefore, in many image processing problems, one can consider a minimization problem with any distribution if the statistical features of the underlying process are known, e.g. Poisson for a counting process, Rayleigh for ultrasound images, and so on. 

This paper is organized as follows: the region-based variational framework for segmentation is introduced in Section \ref{sec:II}. Influence of the parameters estimator is investigated in Section \ref{sec:III}. Our new theoretical results are detailed in Section \ref{sec:IV}. Experimental results are discussed in Section \ref{sec:V}. We finally conclude and give some perspectives.


\section{Problem Statement}
\label{sec:II}

Let $\U$ be a class of domains (open, regular bounded sets, i.e. $C^2$) of $\R^k$, and $\Omega_i$ an element of $\U$ of boundary $\partial\Omega_i$. The image domain is noted $\Om_I$.
A region-based segmentation problem aims at finding a partition of $\Om_I$ in $n$ regions $\{\Om_{1},..,\Om_{n}\}$ of respective boundaries $\{\partial{\Om}_{1},..,\partial{\Om}_{n}\}$ that minimizes the following criterion:
\begin{equation}
\label{eq:global_criterion}
E(\Om_{1},..,\Om_{n},\Gamma) = \sum_{i=1}^{n} E_{ri}(\Om_{i})+\lambda E_b(\Gamma)
\end{equation}
where $E_{ri}$ describes the homogeneity of the region $\Om_{i}$, and $\Gamma=\bigcup_{i=1}^{n}\partial{\Om}_{i}$. 
The energy term $E_b$ is a regularization term balanced with a positive real parameter $\lambda$. 

\subsection{Statistical energy}

Let us denote $\Y(\x)$ a vector of random variables (VRV) at location $\x$ where $\x\in \Om_I$. A region $\Om_i$ is considered to be homogeneous according to the image feature $\y(\x) \in \chi \subset \R^d$ if the associated VRV $\Y(\x)$ follows a prespecified probability distribution with pdf $p(\Y(\x),\vtheta_i)$ of parameters $\vtheta_i$ in the region $\Om_i$. The image feature may be for example the intensity of the region (a 3-dimensional vector for color regions $d=3$, or a scalar for grey level regions $d=1$). We then consider functions of the form:
\begin{equation}
\label{eq:Eri2}
E_{ri}(\Om_{i})=\int_{\Om_i}\Phi( p(\y(\x),\vtheta_i)d\x
\end{equation}
with $\Phi$ at least $C^1$ and Lebesgue integrable function.

Since the above integrals are with respect to domains, and also that the parameters $\vtheta_i$ are region-dependent, optimization (and hence derivation) of (\ref{eq:Eri2}) is not straightforward. The derivation of such a functional is performed using domain derivation tools as in \cite{Delfour_book_01}. 
The following theorem from \cite{Delfour_book_01} will be useful for region-based terms derivation: 
\begin{theorem}
\label{th:der}
The G\^ateaux derivative of the functional \\$J(\Omega)=\int\limits_\Omega
f(\x,\Omega)\,d\x$ in the direction of $\V$ is the following: 
\[
<J'(\Omega),\V>=\int\limits_{\Omega}
f_s(\x,\Omega,\V)d\x-\int\limits_{\partial \Omega} f(\x,\Omega) ( \V \cdot
\boldsymbol{N})  d{\mathbf{a}}(\x)
\]
where $\boldsymbol{N}$ is the unit inward normal to $\partial\Omega$, $d{\mathbf{a}}$ its area element and $f_s$ the shape derivative of $f$.
\end{theorem}

\subsection{Towards a geometrical PDE}

From the shape derivative, we can deduce the evolution equation that will drive the active contour towards a minimum of the criterion.
In order to fix ideas, let us consider the case of a partition of an image in two regions $\{\Om_1,\Om_2\}$. The curve $\Gamma$ stands for the interface between the two regions. The boundary energy term $E_b$ may be chosen as the curve length and derived classically using calculus of variation \cite{Caselles_ijcv_97} or shape derivation tools \cite{Delfour_book_01}.

Let us suppose that the shape derivative of each region $\Om_i$ may be written as follows:
\begin{equation}
\label{eq:Eul_der_ri}
<E_{ri}'(\Omega_i),\V>=-\int\limits_{\partial \Omega_i} f_i(\x,\Om_i) ( \V(\x) \cdot \boldsymbol{N(\x)})  d\mathbf{a}(\x) 
\end{equation}





In this paper, we focus our attention on the computation of the shape derivative of the criterion (\ref{eq:Eri2}). Let us note that when $\Phi(t)=-\log(t)$ the function (\ref{eq:Eri2}) is known as the log-likelihood score function, used to describe the homogeneity of a region according to a prescribed noise model. It has been used in \cite{Zhu_pami_96,Paragios_jvcir_02,Refregier_ip04}. Under some conditions, we are able to prove that minimizing this function with the ML estimator gives a simple evolution speed whose closed-form expression depends only on the pdf. However, one must be aware that using a more general function $\Phi$ or different hyperparameter estimator will generally yield complicating additive terms in the shape derivative and hence in the evolution speed expression.

\section{Influence of the estimator on the evolution equation}
\label{sec:III}

This section is devoted to illustrate two examples that support our previous claims on the influence of the parameters estimator. More precisely, we state two derivative results in the case of the Rayleigh distribution with two different estimators (ML and moments).

\subsection{Shape derivative for the Rayleigh distribution}

We consider the Rayleigh distribution of scalar parameter $\theta_i$:
\begin{equation}
p(\y(\x),\theta_i)=\frac{\y(\x)}{\theta_i^2}\exp{\left(\frac{-\y(\x)^2)}{2\theta_i^2}\right)}
\end{equation}
Classically, one can compute an estimate of the parameter $\theta_i$ using the moment method. In this case the estimator is given by :
\begin{equation}
\label{eq:estmo}
\ESmo=\sqrt{\frac{2}{\pi}}\ \frac{1}{|\Om_i|}\int_{\Om_i} \y(\x)d\x =\sqrt{\frac{2}{\pi}} \ \overline{\y(\x)}_i
\end{equation}
where $|\Om_i|=\int_{\Om_i}d\x$ and $\overline{\y(\x)}_i$ denotes the sample mean inside the region $\Om_i$.\\
Alternatively, one can also compute an estimate using the ML estimator given by:
\begin{equation}
\label{eqmv}
\ESmv=\sqrt{\frac{1}{2 |\Om_i|}\int_{\Om_i} \y(\x)^2 d\x }
\end{equation} 

\subsubsection{Shape derivative with the moment estimator}
\begin{theorem}
\label{th:der_ray_mo}
The G\^ateaux derivative, in the direction of $\V$, of the functional $E_{ri}(\Om_i)=-\int_{\Om_i} \log( p(\y(\x),\ESmo)d{\mathbf{a}}(\x)$ with $p$ a Rayleigh distribution, is the following: 
\begin{eqnarray}
<E_{ri}'(\Om_i),\V>&=&\int_{\partial{\Om_i}}(\log{(p(\y(\x),\ESmo)}\nonumber\\
&+&A(\y(\x),\Om_i)( \V \cdot\boldsymbol{N})  d{\mathbf{a}}(\x)\nonumber
\end{eqnarray}
\[
A(\y(x),\Om_i)=\Big(2-\frac{\pi}{4}\frac{\overline{\y^2(\x)}_i}{\overline{\y(\x)}_i^2}\Big)\Big(1-\frac{\y(\x)}{\overline{\y(\x)}_i}\Big)
\]
where $\overline{\y^k(\x)}_i=\frac{1}{|\Om_i|}\int_{\Om_i} \y^k(\x) d\x$.

\end{theorem}

\subsubsection{Shape derivative with the ML estimator}

\begin{theorem}
\label{th:der_ray_mv}
The G\^ateaux derivative, in the direction of $\V$, of the functional $E_{ri}(\Om_i)=-\int_{\Om_i} \log( p(\y(x),\ESmv)d\x$ with $p$ a Rayleigh distribution, is the following: 
\[
<E_{ri}'(\Om_i),\V>=\int_{\partial{\Om_i}} \log{\left(p(\y(\x),\ESmv)\right)} ( \V \cdot \N)  d{\mathbf{a}}(\x)
\]
\end{theorem}

One can then see that the parameter estimator has a clear impact on the evolution speed expression, as the additive term $A(\y(x),\Om_i)$ appears in the shape derivative when the moment estimator is used.

\section{General results for the exponential family}
\label{sec:IV}

Let us now consider the shape derivative within the framework of multi-parameter exponential family. This family includes Poisson, Rayleigh, Gaussian... 

\subsection{Exponential families}

\subsubsection{Definition}
The multi-parameter exponential families are naturally indexed by a $k$-real parameter vector and a $k$-dimensional natural statistics $T(\Y)$. A simple example is the normal family when both the location and the scale parameters are unknown ($k=2$).

\begin{definition}
The family of distributions of a VRV $\Y$ $\{\mathcal{P}_\veta: \veta \in \mathcal{E} \subseteq \R^k\}$, is said a k-{\em{parameter {\bf canonical} exponential family}}, if there exist real-valued functions $\eta_1, \ldots, \eta_k: \Theta \mapsto \R$ and $A(\veta)$ on $\mathcal{E}$, and real-valued functions $h, T_1, \ldots, T_k: \R^d \mapsto \R$, such the pdf $p(\y,\veta)$ of the $\mathcal{P}_\veta$ may be written:
\begin{equation}
\label{eqdef3}
p(\y,\veta) = h(\y) \exp [\langle\veta,\vT(\y)\rangle - A(\veta)], \y \in \chi \subset \R^d
\end{equation}
where $\vT=(T_1,\ldots,T_k)^T$ is the {\em{natural sufficient statistic}}, $\veta=(\eta_1,\ldots,\eta_k)^T$ and $\mathcal{E}$ are the {\em{natural parameter vector and space}}. $\langle\veta,\vT\rangle$ denotes the scalar product.
\end{definition}
We draw the reader's attention to the fact than $\veta$ is a function of $\theta$ which is the parameter of interest in most applications.

For example, for the Rayleigh distribution : $\veta=-\frac{1}{2\theta^2}$, $T(\y)=\y^2$, $h(\y)=\frac{\y}{\theta^2}$ and $A(\veta)=-\log(-2\veta)$.

\subsubsection{Properties}
The following theorem establishes the conditions of strict convexity of $A$, and then those for $\nabla A$ to be 1-1 on $\mathcal{E}$. This is very useful result for optimization (derivation) purposes:
\begin{theorem}
\label{theo3}
Let $\mathcal{P}$ a full rank (i.e. $Cov[T(\Y)]$ is a positive-definite matrix) $k$-parameter canonical exponential family with natural sufficient statistic $\vT(\Y)$ and open natural parameter space $\mathcal{E}$ \cite{BickelBook2001}. 
\begin{enumerate}[(i)]
\item \label{iit3} $\nabla A: \mathcal{E} \mapsto \mathcal{S} \subseteq \R$ is 1-1. The family may be uniquely parameterized by $\vmu(\veta) \equiv E(\vT(\Y)) = \nabla A(\veta)$.
\item The -log-likelihood function is a strictly convex function of $\veta$ on $\mathcal{E}$.
\end{enumerate}
\end{theorem}

These results establish a 1-1 correspondence between $\veta$ and $E(\vT(\Y))$ such that:
\begin{equation}
\mathcal{S} \ni \vmu = \nabla A(\veta) = E(\vT(\Y)) \Leftrightarrow \mathcal{E} \ni \veta = \psi\left(E(\vT(\Y))\right)
\end{equation}
holds uniquely with $\nabla A$ and $\psi$ continuous. 

\subsection{General results}

In the sequel, for the sake of simplicity, we will invariably denote $\veta$ for the natural parameter and its finite sample estimate over the domain (without a slight abuse of notation, this should be $\hat{\veta}$). 
\begin{theorem}
\label{th:gene}
The G\^ateaux derivative, in the direction of $\V$, of the functional $E_{ri}(\Om_i)=\int_{\Om_i} \Phi(p(\y(\x),\veta(\Om_i)))d{\mathbf{a}}(\x)$ where $p(.)$ belongs to the multi-parameter exponential family and is expressed by (\ref{eqdef3}) and $\veta$ the natural hyperparameter vector, is:
\begin{eqnarray}
\label{derE}
<E_{ri}'(\Om_i),\V>= -\int_{\partial\Omega}{\Phi(p(\y))(\V \cdot \N) d{\mathbf{a}}(\x)} \nonumber
\\ + \int_{\Om_i} p(\y) \Phi'(p(\y))\langle\nabla_{\V}\veta,\T(\y)-\nabla A(\veta)\rangle d\x
\end{eqnarray}
with $\nabla_{\V}\veta$ the G\^ateaux derivative of $\veta$ in the direction of $\V$, and $\langle \x,\y \rangle$ the scalar product of vectors $\x$ and $\y$.
\end{theorem}


In a finite sample setting, when using the ML estimator, we can replace $\nabla A(\veta)$ by $\overline{\T(\Y)}$ (the 1st order sample moment of $\T(\Y))$. Thus, when using the -log-likelihood function, the second term becomes equal to $\int_{\Om_i}\langle\nabla_{\V}\veta,\T(\y)-\overline{\T(\Y)})\rangle d\x$, and hence vanishes. The following corollary follows:

\begin{corollaire}
\label{th:der_multipar}
The G\^ateaux derivative, in the direction of $\V$, of the functional $E_{ri}(\Om_i)=-\int_{\Om_i} \log(p(\y(\x),\mathbf{\vetaML}(\Om_i))d{\mathbf{a}}(\x)$ when $\mathbf{\vetaML}$ is the ML estimate, is the following: 
\[
<E_{ri}'(\Om_i),\V>=\int_{\partial{\Om_i}}(\log{(p(\y(\x),\mathbf{\vetaML}(\Om_i)))}( \V \cdot\boldsymbol{N})  d{\mathbf{a}}(\x)
\]
\end{corollaire}

This provides an alternative proof to the result of \cite{Zhu_pami_96,Refregier_ip04}. Nonetheless, we here point out that, in the work of \cite{Zhu_pami_96,Refregier_ip04}, the role of the parameters estimator was not elucidated.

\section{Experimental Results}
\label{sec:V}
\begin{figure}
\center
\includegraphics[width=0.33\linewidth]{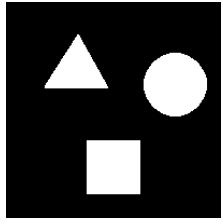}
\caption{Initial noise-free image.}
\label{fig:init}
\end{figure}

This section presents some experimental results on noisy images. The initial noise-free image is shown in Fig.\ref{fig:init}. For four different Battacharya distances (BD), we have systematically corrupted this image with two types of noise: Poisson and Rayleigh. The Battacharya distance is used as a measure of "contrast" between objects and background. It is defined as : $$\mathcal{D}(p_f(\y),p_o(\y))=-\log \int_{\R^d} \sqrt{p_f(\y)p_o(\y)}d\y$$ 
For each combination of BD value and noise type, 50 noisy images were generated. Each noisy image was then segmented using four different energy functionals, namely Chan-Vese \cite{Vese_ijcv02}, and our method with -log-likelihood and ML estimator with three assumed noise models: Gaussian, Rayleigh and Poisson. For each segmented image with each method at each BD value, the average false positive fraction (FPF) and true positive fraction (TPF), over the 50 simulations were computed. The bottomline of these experiments is to show that using the appropriate noise model will yield the best performance in terms of compromise between specificity (over-segmentation as revealed by the FPF) and sensitivity (under-segmentation as revealed by the TPF).

\begin{figure}
\center
\includegraphics[width=\linewidth]{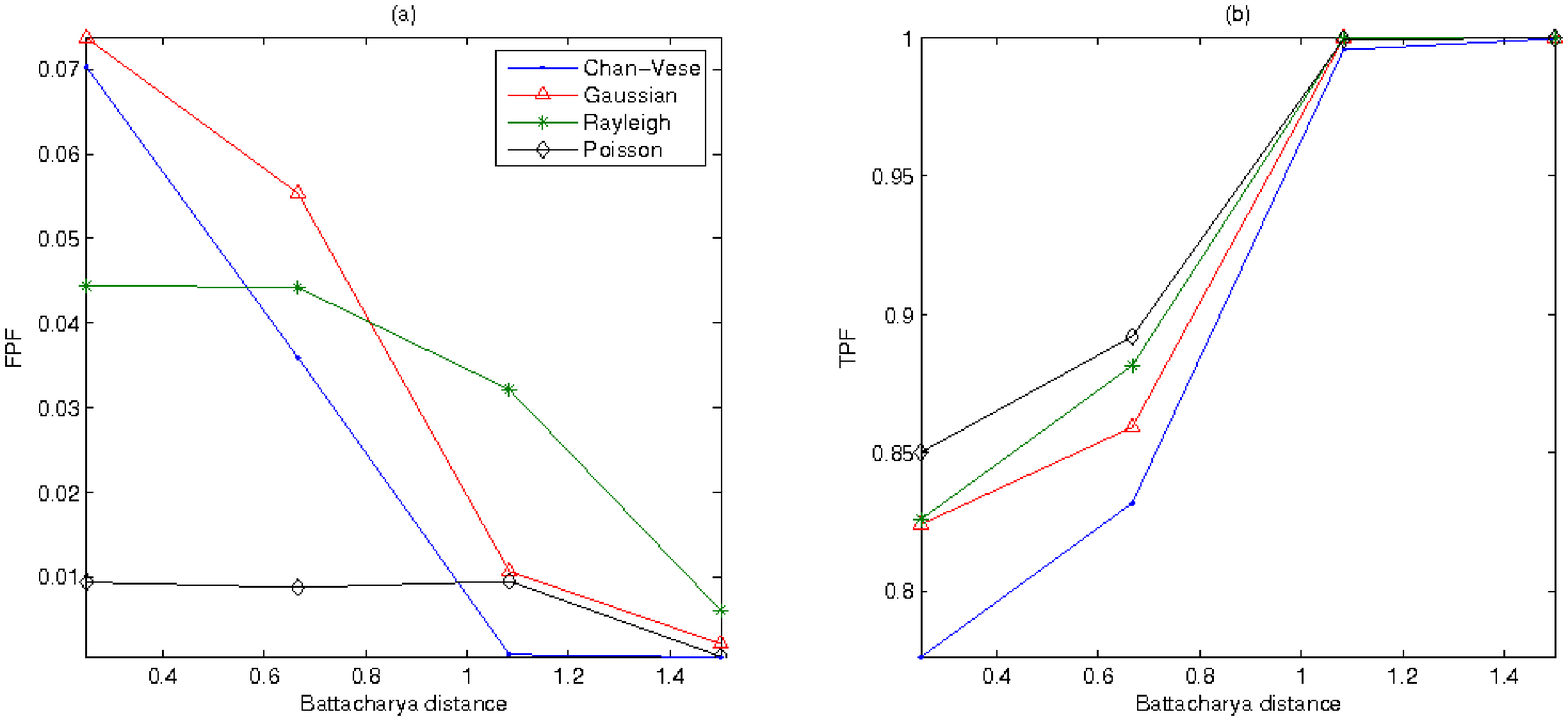}\\
\includegraphics[width=\linewidth]{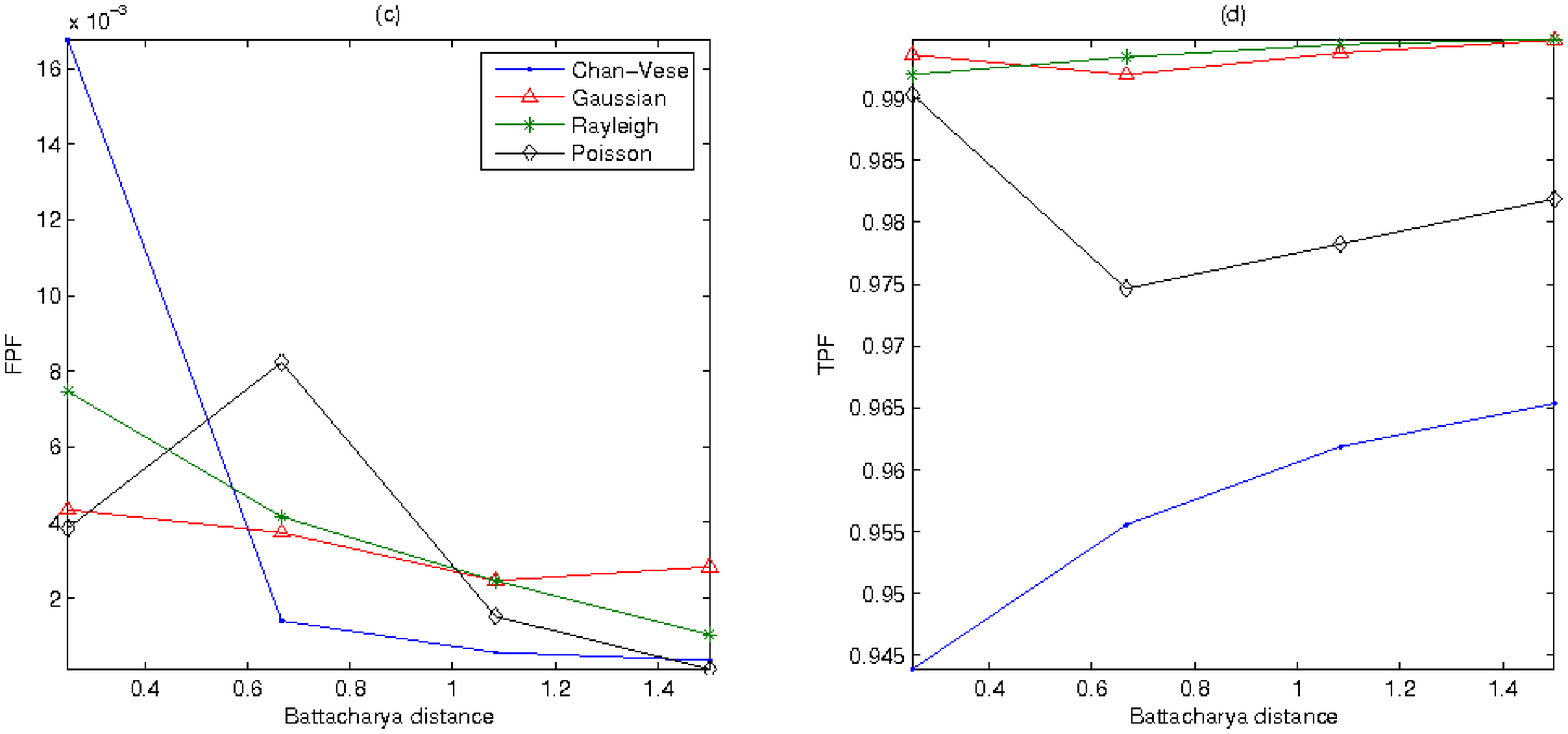}
\caption{FPF and TPF as a function of BD.}
\label{fig:all}
\end{figure}

Fig.\ref{fig:all} depicts the average FPF (left) and TPF (right) as a function of the BD for Poisson ((a)-(b)) and Rayleigh ((c)-(d)) noises. As expected, the FPF exhibits a decreasing tendency as the BD increases, while the TPF increases with BD, which is intuitively acceptable. More interestingly, the best performance in terms of compromise between FPF and TPF is reached when the contaminating noise and the noise model in the functional are the same. This behaviour is more salient at low BD values, i.e. high noise level. One can also point out that the Chan-Vese functional is very conservative at the price of less sensitivity. Clearly this method under-segments the objects.


\vspace{-0.5cm}
\section{Conclusion}
\label{sec:VI}
In this work, we proposed a novel statistical region-based active contours method, where the region descriptor is written as a function $\Phi$ of some pdf belonging to the exponential family. The case of the likelihood score, which describes the homogeneity of the regions, is obtained as a special case of our setting. We shedded light on the influence of the noise parameters estimator, and on the adequacy between the noise in the observations and the one incorporated in the active contours functional. Our ongoing work is directed towards extending our approach to $n$ regions by adapting the multiphase method \cite{Vese_ijcv02}. Furthermore, encouraged by our preliminary application results, a deeper experimental work is currently carried out to validate our approach on real data sets.

\small{
\bibliographystyle{IEEEbib}
\bibliography{thesis}
}

\end{document}